\documentclass[rmp,aps,twocolumn,nofootinbib]{revtex4}
\usepackage{pslatex} % font: pslatex, helvet, palatino, times
\usepackage{amsmath}
\usepackage[]{graphicx} % graphics for figures

\long\def\symbolfootnote[#1]#2{\begingroup \def\thefootnote{\fnsymbol{footnote}}\footnote[#1]{#2}\endgroup}
\newcommand{\comment}[1]{} % block comment

\newcommand{\argmin}[1]{\operatorname*{arg\,min}_#1}

\begin{document}

\title{A Tutorial on Independent Component Analysis}
\date{\today; Version 1.0}
\author{Jonathon Shlens} 
\email{jonathon.shlens@gmail.com}
\affiliation{
Google\;Research\\
Mountain View, CA  94043}

\begin{abstract}
Independent component analysis (ICA) has become a standard data analysis technique applied to an array of problems in signal processing and machine learning. This tutorial provides an introduction to ICA based on linear algebra formulating an intuition for ICA from first principles. The goal of this tutorial is to provide a solid foundation on this advanced topic so that one might learn the motivation behind ICA, learn why and when to apply this technique and in the process gain an introduction to this exciting field of active research.
\end{abstract}

\maketitle

\section{Introduction}

Measurements often do not reflect the very thing intended to be measured. Measurements are corrupted by random noise -- but that is only one piece of the story. Often, measurements can not be made in isolation, but reflect the combination of many distinct sources. For instance, try to record a person's voice on a city street. The faint crackle of the tape can be heard in the recording but so are the sounds of cars, other pedestrians, footsteps, etc. Sometimes the main obstacle preventing a clean measurement is not just noise in the traditional sense (e.g. faint crackle) but independent signals arising from distinct, identifiable sources (e.g. cars, footsteps).

The distinction is subtle. We could view a measurement as an estimate of a single source corrupted by some random fluctuations (e.g. additive white noise). Instead, we assert that a measurement can be a combination of many distinct sources -- each different from random noise. The broad topic of separating mixed sources has a name - {\it blind source separation} (BSS). As of today's writing, solving an arbitrary BSS problem is often intractable. However,  a small subset of these types of problem have been solved only as recently as the last two decades -- this is the provenance of {\it independent component analysis} (ICA).

Solving blind source separation using ICA has two related interpretations -- filtering and dimensional reduction. If each source can be identified, a practitioner might choose to selectively delete or retain a single source (e.g. a person's voice, above). This is a filtering operation in the sense that some aspect of the data is selectively removed or retained. A filtering operation is equivalent to projecting out some aspect (or dimension) of the data -- in other words a prescription for dimensional reduction. Filtering data based on ICA has found many applications including the analysis of photographic images, medical signals (e.g. EEG, MEG, MRI, etc.), biological assays (e.g. micro-arrays, gene chips, etc.) and most notably audio signal processing.

ICA can be applied to data in a naive manner treating the technique as a sophisticated black box that essentially performs ``magic''. While empowering, deploying a technique in this manner is fraught with peril. For instance, how does one judge the success of ICA? When will ICA fail? When are other methods more appropriate? I believe that understanding these questions and the method itself are necessary for appreciating when and how to apply ICA. It is for these reasons that I write this tutorial.

This tutorial is not a scholarly paper. Nor is it thorough. The goal of this paper is simply to educate. That said, the ideas in this tutorial are sophisticated. I presume that the reader is comfortable with linear algebra, basic probability and statistics as well as the topic of principal component analysis (PCA).\footnote{A good introduction to PCA written in the same style is {\it A Tutorial on Principal Component Analysis} by yours truly. This tutorial provides a full introduction and discussion of PCA with concrete examples for building intuition.} This paper does not shy away from informal explanations but also stresses the mathematics when they shed insight on to the problem. As always, please feel free to email me with any comments, concerns or corrections.\footnote{I wish to offer a special thanks to E. Simoncelli for many fruitful discussions and insights and U. Rajashekar for thoughtful discussion, editing and feedback.}

\section{Measurements in the Real World}
While the mathematics underlying the problems of blind source separation (BSS) are formidable, the ideas underlying BSS are quite intuitive. This section focuses on building this intuition by asking how signals interact in the real world when we make a measurement in which multiple sources contribute in addition to noise.

Many experiments have domain-specific knowledge behind how a measurement is performed. (How does a gene chip work?) Instead, this tutorial describes how measurements are made in domains in which we are all experts -- our own senses. We discuss what a BSS problem entails in the visual and auditory domains and in the process build some intuition for what it takes to solve such a problem.

\begin{figure}[t]
\vspace{5pt}
\centering
\includegraphics[width=0.47\textwidth]{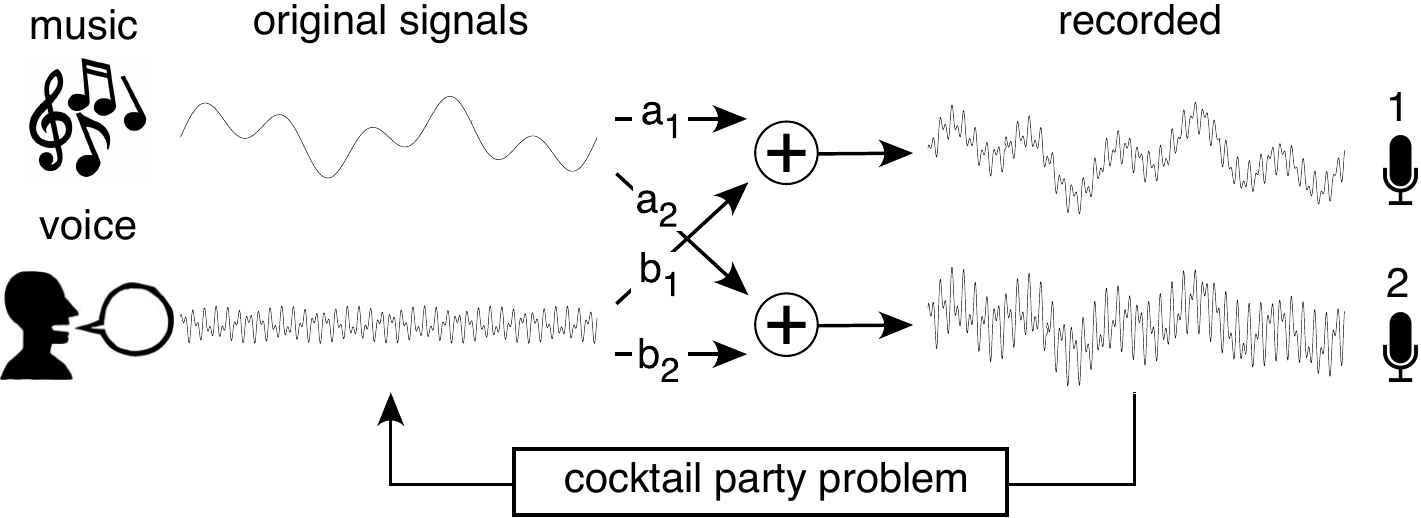}
\caption{Example of the cocktail party problem. Two sounds $s_1$, $s_2$ are generated by music and a voice and recorded simultaneously in two microphones. Sound adds linearly. Two microphones record a unique linear summation of the two sounds. The linear weights for each microphone ($a_1$, $b_1$ and $a_2$, $b_2$) reflect the proximity of each speaker to the respective microphones. The goal of the cocktail party problem is to recover the original sources (i.e. music and voice) solely using the microphone recordings \cite{Bregman94}.}
\label{fig:cocktail}
\vspace{15pt}
\end{figure}

No better example exists of a BSS problem than sound. A fundamental observation from physics is that sound adds linearly. Whatever is recorded in a microphone or registered in our ear drum is the sum of the pressure waves from emanating from multiple sources. Consider an everyday problem. Amongst a background of music and conversations, we must discern a single person's voice. The {\it cocktail party problem} is a classic problem in auditory signal processing. The goal of the cocktail party problem is to discern the sound associated with a single object even though all of the sounds in the environment are superimposed on one another (Figure~\ref{fig:cocktail}). This is a challenging and highly applicable problem that only in recent years have reasonable solutions been devised based on ideas from ICA.

\begin{figure}[h]
\centering
\includegraphics[width=0.47\textwidth]{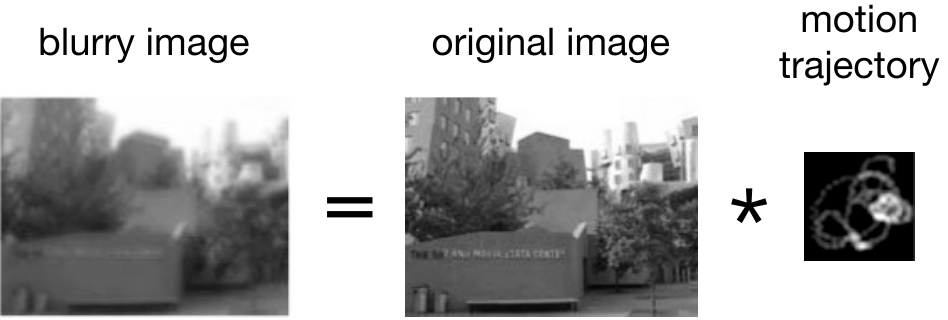}
\caption{Example of removing blur from an image due to camera motion. A blurry image (left panel) recorded on a camera sensory array is approximately equal to the convolution of the original image (middle panel) and the motion path of the camera (right panel). Each pixel in the blurry image is the weighted sum of pixels in the original image along the camera motion trajectory. De-blurring an image requires identifying the original image and the motion path from a single blurry image (reproduced from \textcite{Fergus06}).}
\label{fig:blur}
\vspace{15pt}
\end{figure}

Image processing provides challenging BSS problems as well. Consider the problem of removing blur from an image due to camera motion (Figure~\ref{fig:blur}). A photographer tries to take a photo but their camera is not steady when the aperture is open. Each pixel in the sensor array records the sum of all light within an integration period from the intended image along the camera motion trajectory. Each point in the recorded image can be viewed as the temporally weighted sum of light from the original image along the camera motion trajectory. Thus, each recorded pixel is the linear mixture of multiple image pixels from the original image. De-blurring an image requires recovering the original image as well as the underlying camera motion trajectory from the blurry image \cite{Fergus06}. Both the cocktail party and de-blurring problems are ill-posed and additional information must be employed in order to recover a solution.

These examples from vision and audition highlight the variety of signal interactions in the real world. No doubt more complex interactions exist in other domains or in measurements from specific instruments. Recovering individual sources from combined measurements is the purview of BSS and problems where the interactions are arbitrarily complex, are generally not possible to solve.

That said, progress has been made when the interactions between signals are simple -- in particular, {\it linear} interactions, as in both of these examples. When the combination of two signals results in the superposition of signals, we term this problem a linear mixture problem. {\it The goal of ICA is to solve BSS problems which arise from a linear mixture.} In fact, the prototypical problem addressed by ICA is the cocktail party problem from Figure~\ref{fig:cocktail}. Before diving into the mathematics of ICA it is important to visualize what linear mixed data ``looks like'' to build an intuition for a solution.

\begin{figure}[!]
\centering
\includegraphics[width=0.47\textwidth]{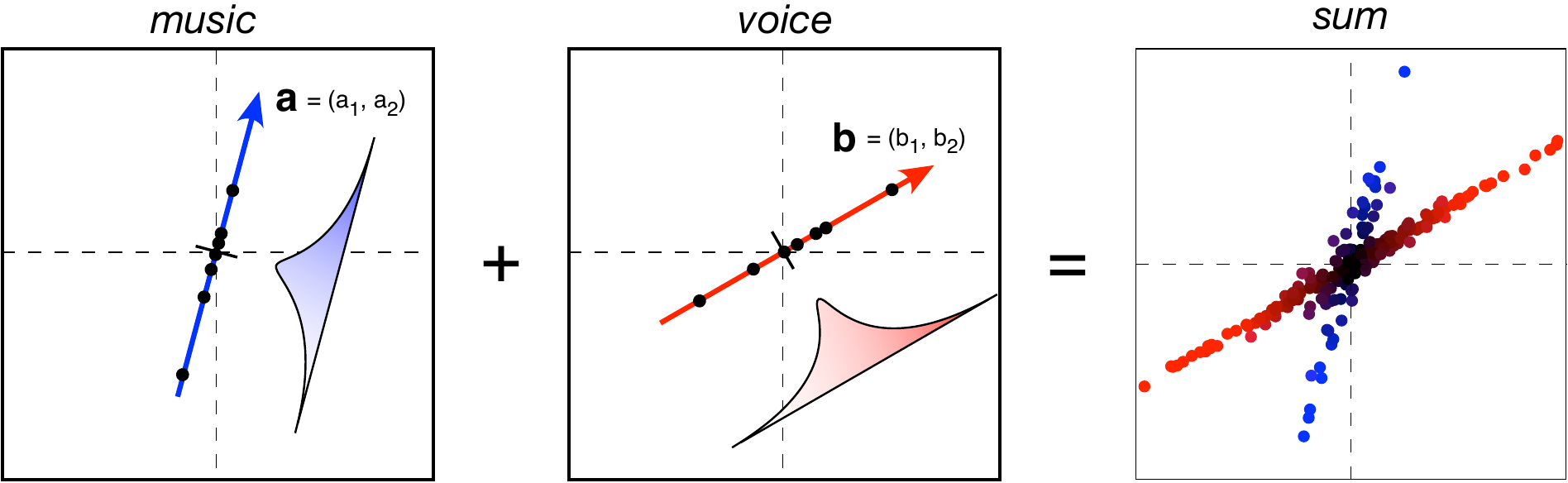}
\caption{Example data from the cocktail party problem. Amplitudes recorded simultaneously in microphones 1 and 2 are plotted on the $x$ and $y$ axes, respectively. In the left panel all samples arising from the music lie on the vector $\mathbf{a}=(a_1, a_2)$ reflecting the proximity of the music to microphones 1 and 2, respectively. Likewise, the middle panel depicts all data points arising from solely the voice. The right panel depicts recording of the linear sum both sources. To highlight the contribution of the voice and the music we color each recorded sample by the relative contribution of each each source.}
\label{fig:cocktail-data}
\vspace{15pt}
\end{figure}

\section{Examples on Linear Mixed Signals}

\begin{figure}[t]
\centering
\includegraphics[width=0.47\textwidth]{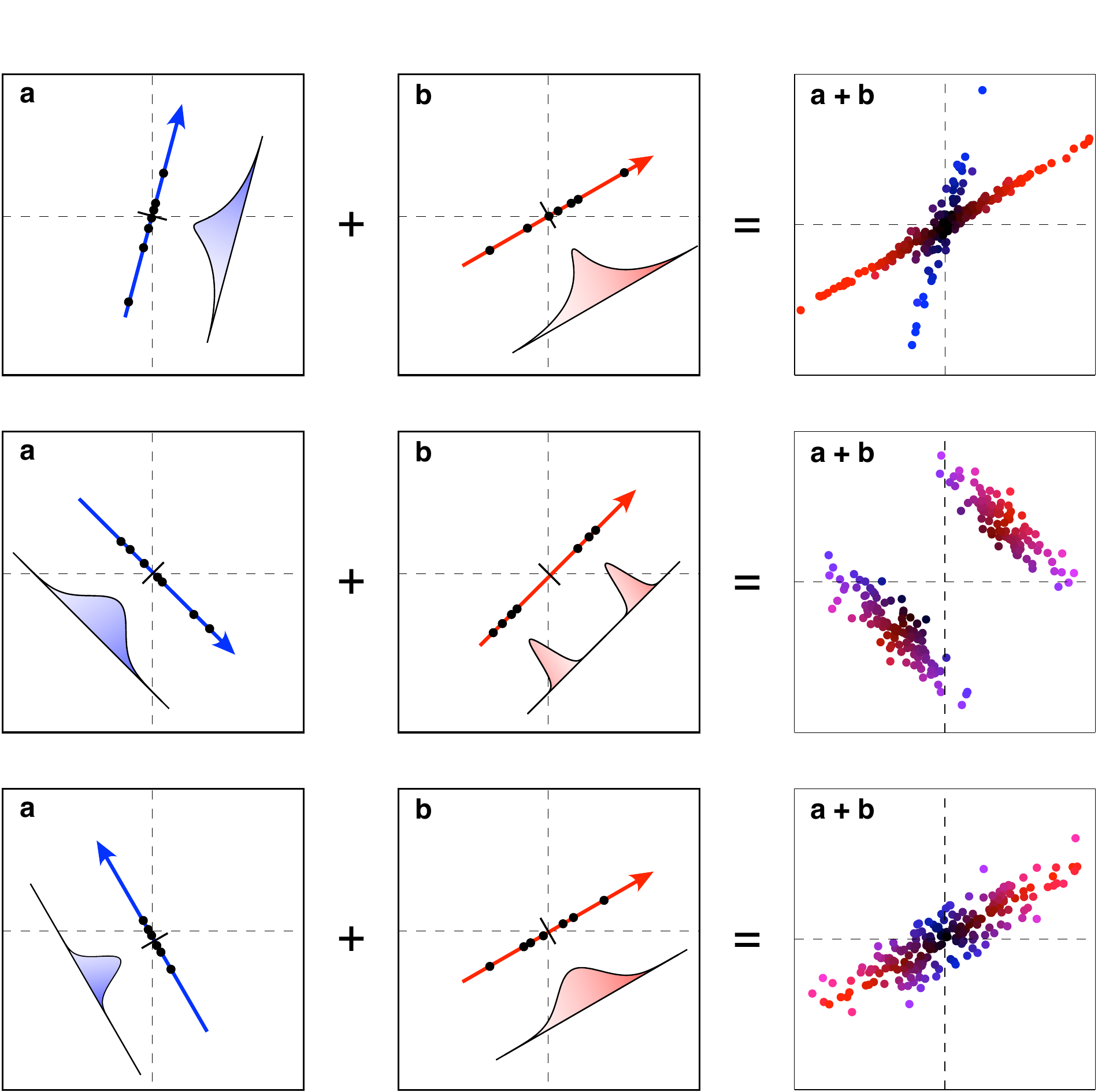}
\caption{Examples of linear mixed data. Consider two sources $\mathbf{a}$ and $\mathbf{b}$ represented as blue and red vectors ($n=2$ dimensions). Each source has a direction represented by the vector termed the {\em independent components} and a magnitude whose amplitude varies (randomly) according to some distribution. The sum of the independent components weighted by each sample's magnitude $\mathbf{a + b}$ produces a data point whose color represents the sum of the magnitudes of the sources which contributed to the data point. {\em Top row.} The combination of two sharply peaked sources produces an {\tt X} formation. {\em Middle row.} Same as top row but the first and second sources are unimodal and bimodal Gaussian distributed (adapted from A. Gretton). {\em Bottom row.} Same as top but both sources are Gaussian distributed.}
\label{fig:linear-mixing}
\end{figure}

\begin{figure}[t]
\centering
\includegraphics[width=0.47\textwidth]{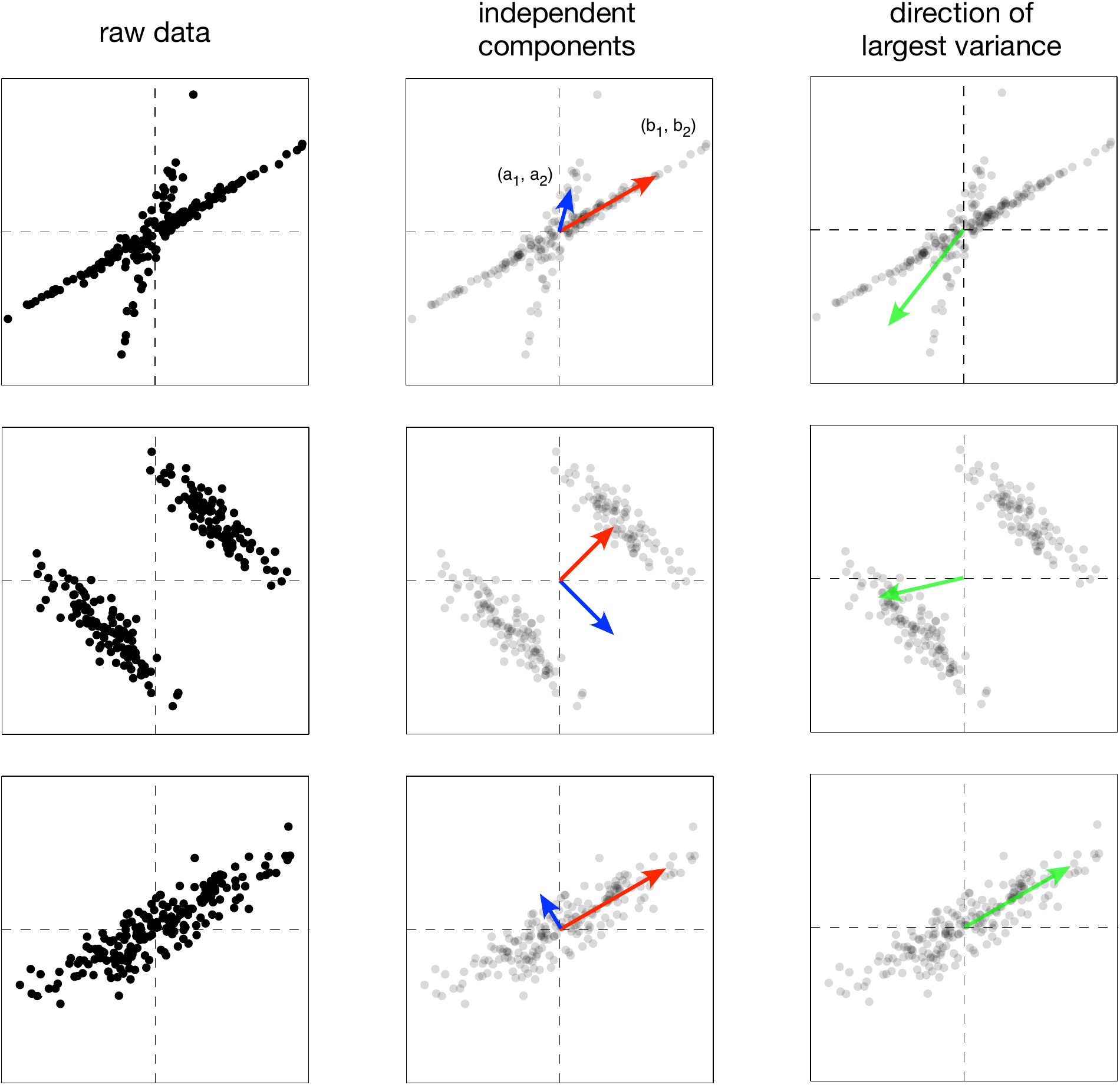}
\caption{Analysis of linear mixed signals. {\em Left column.} Data is reproduced from Figure~\ref{fig:linear-mixing}. {\em Middle column.} Data is replotted superimposed on the basis of the underlying sources, i.e. the independent components (red and blue arrows) in the color corresponding to the basis used to generate the data in Figure~\ref{fig:linear-mixing}. {\em Right column.} Data is replotted superimposed on the direction of largest variance (green arrows). Note that the direction of largest variance might or might not correspond to the independent components.}
\label{fig:examples}
\end{figure}

Let us imagine data generated from the cocktail party problem in Figure~\ref{fig:cocktail}. Microphone 1 records the music and voice with proximities $a_1$ and $b_1$, respectively. We plot on the $x$ and $y$ axes the recorded values for microphones 1 and 2, respectively in Figure~\ref{fig:cocktail-data}.

What happens if music alone is played? What would the data look like?  If music alone is played without a voice, then any data point would lie along the vector $(a_1, a_2)$. Why? The reason is that $(a_1, a_2)$ measures the proximity of the music to microphones 1 and 2, respectively. Since the music resides at a fixed position with respect to the microphones, each recorded data point must be yoked to $(a_1, a_2)$ as the volume of the music varies over time. Finally, all of the music volumes across time are sampled from some underlying distribution. This behavior is depicted the left panel of Figure~\ref{fig:cocktail-data}.

The same experiment could be applied to the data arising from only the voice. This is depicted in the middle panel of Figure~\ref{fig:cocktail-data}. Note that the bases associated with each audio source $(a_1, a_2)$ and $(b_1, b_2)$ merely reflect the proximity of each audio source to the microphones. Therefore, the bases need not be orthogonal on the $(x,y)$ plot. Each basis vector labeled $\mathbf{a}=(a_1, a_2)$ and $\mathbf{b}=(b_1, b_2)$, respectively, is termed the independent component (IC) of the data.

Finally, If both the music and voice are played together, the simultaneous measurement in microphones 1 and 2 would be the vector sum of the samples drawn from each basis (again, because sound adds linearly). The composite recording is depicted in the top right panel of Figure~\ref{fig:linear-mixing}, where each data point is colored according to the relative contribution of the music and voice (red and blue, respectively).

The middle and bottom rows of Figure~\ref{fig:linear-mixing} depict two different examples of linear mixed data to give a sense of the diversity of BSS problem. In the middle row, the two ICs are orthogonal but the distributions along each axis are dramatically different -- namely unimodal and bimodal distributions, respectively. The resulting distribution appears as two oriented lobes. The bottom row depicts two orthogonal ICs, each Gaussian distributed. The resulting distribution is also Gaussian distributed.

In the situation confronted by an experimentalist, the observed data contain no ``color labeling'' and all analyses must be performed solely on the linear sum (Figure~\ref{fig:examples}, left column). Examining the black points in each example, an experimenter might wish to select a ``piece'' of the data -- for instance, one arm of the {\tt X}\,, one mode of the bimodal data or one direction of the Gaussian distribution in each panel respectively. Because this author judiciously constructed these examples, it is deceptively easy to filter out the distracting source and extract the desired source from each example (e.g. music or voice). Note that these examples are two-dimensional and although we have the benefit of {\it seeing} the answer, in higher dimensions visualization becomes difficult and a solution might not be as salient.

Because a solution is salient in these examples, it is worthwhile to codify the solution in order to understand what we are aiming for in more complex data sets. The solution to each example is to identify the basis underlying each source, i.e. the independent components. This is depicted by the red and blue arrows in the middle column of Figure~\ref{fig:examples}. The independent components (ICs) might not be orthogonal (top row). Furthermore, the ICs often do not correspond with the direction of maximal variance (top and middle rows). Note that the latter points imply that any technique simply examining the variance would fail to find the ICs (although quizzically the variance does seem to identify one IC in the bottom example). Understanding these issues and how to recover these bases are the heart of this tutorial.

The goal of this tutorial is to build a mathematical solution to the intuition highlighted in all of these examples. We term the solution ICA. During the course of the manuscript, we will understand how and why the types of distributions play an intimate role in this problem and understand how to recover the independent components of most any data set.

\section{Setup\label{sec:setup}}

Here is the framework. We record some multi-dimensional data $\mathbf{x}$. We posit that each sample is a random draw from an unknown distribution $P(\mathbf{x})$ (e.g. black points in Figure~\ref{fig:examples}). To keep things interesting, we consider $\mathbf{x}$ to contain more than one dimension.\footnote{A word about notation. Bold lower case letters $\mathbf{x}$ are column vectors whose $i^{th}$ element is $x_i$, while bold upper case letters $\mathbf{A}$ are matrices. The probability of observing a random variable $Y$ is formally $P_Y(Y=y)$ but will be abbreviated $P(y)$ for simplicity.}

We assume that there exists some underlying sources $\mathbf{s}$ where each source $s_i$ is statistically independent -- the observation of each source $s_i$ is independent of all other sources $s_j$ (where $i \neq j$). For instance, in the cocktail party problem the amplitude of the voice $s_1$ is independent of the amplitude of the music $s_2$ at each moment of time.

The key assumption behind ICA is that the observed data $\mathbf{x}$ is a {\em linear} mixture of the underlying sources
\begin{equation}
\mathbf{x} =\mathbf{A}\mathbf{s}
\label{eq:linear-mixture}
\end{equation}
where $\mathbf{A}$ is some unknown invertible, square matrix that mixes the components of the sources. In the example from Figure~\ref{fig:cocktail} $\mathbf{A} = \left[\begin{array}{cc} a_1 & b_1 \\ a_2 & b_2 \end{array}\right]$.The goal of ICA is to find the {\em mixing} matrix $\mathbf{A}$ (more specifically, the inverse of $\mathbf{A}$) in order to recover the original signals $\mathbf{s}$ from the observed data $\mathbf{x}$. 

We will construct a new matrix $\mathbf{W}$ such that the linear transformed data is an estimate of the underlying sources,
\begin{equation}
\mathbf{\hat{s}} = \mathbf{W}\mathbf{x}
\label{eq:ica}
\end{equation}
In this setting the goal of ICA is to find an {\em unmixing} matrix $\mathbf{W}$ that is an approximation of $\mathbf{A}^{-1}$ so that $\mathbf{\hat{s}} \approx \mathbf{s}$.

On the face of it, this might appear an impossible problem: find two unknowns $\mathbf{A}$ and $\mathbf{s}$ by only observing their matrix product $\mathbf{x}$ (i.e. black points in Figure~\ref{fig:examples}). Intuitively, this is akin to solving for $a$ and $b$ by only observing $c = a\times b$ . Mathematically, one might call this an {\em under-constrained} problem because the number of unknowns exceed the number of observations.

This intuition is precisely what makes ICA a challenging problem. What I hope to convey to you is that by examining the statistics of the observed data $\mathbf{x}$, we can find a solution for this problem. This prescription is the essence of ICA.

\section{A Strategy for Solving ICA\label{sec:strategy}}

\begin{figure*}[t]
\centering
\includegraphics[width=0.75\textwidth]{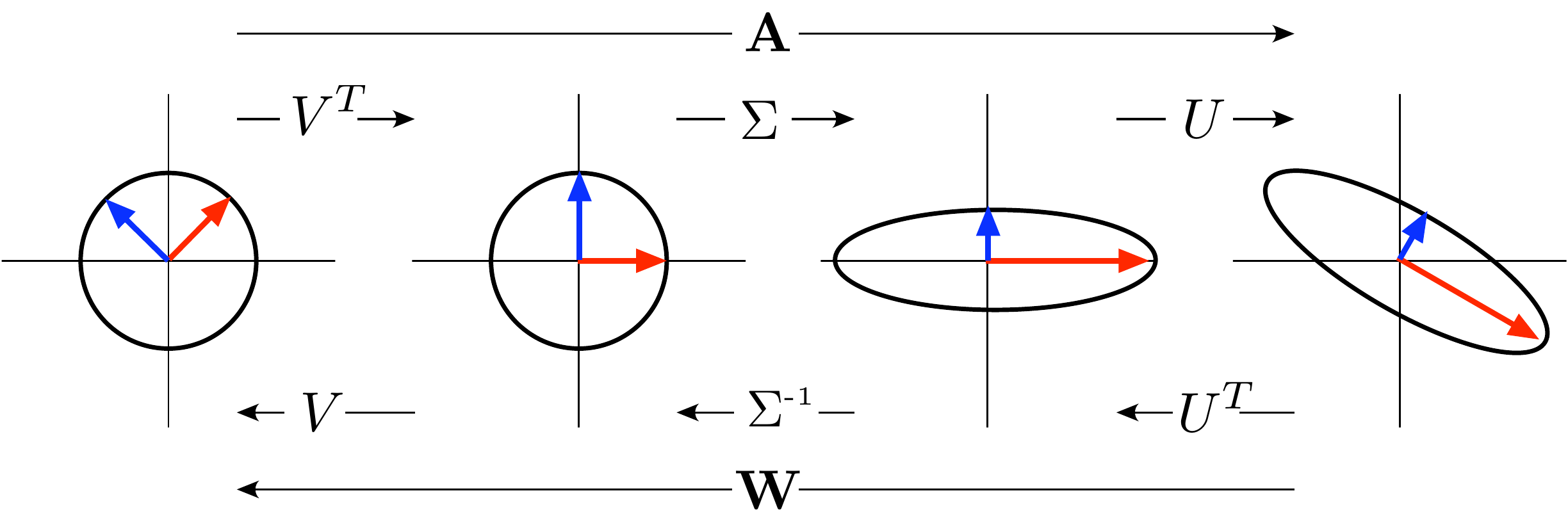}
\caption{Graphical depiction of the singular value decomposition (SVD) of matrix $\mathbf{A} =\mathbf{U}\Sigma\mathbf{V}^T$ assuming $\mathbf{A}$ is invertible. $\mathbf{V}$ and $\mathbf{U}$ are rotation matrices and $\Sigma$ is a diagonal matrix. Red and blue arrows are vectors that correspond to the columns of matrix $\mathbf{V}$ (i.e. the basis of the row space of $\mathbf{A}$). Note how the basis rotates, stretches and rotates during each successive operation. The composition of all three matrix operations is equal to the operation performed by $\mathbf{A}$. The inverse of matrix $\mathbf{A}$ defined as $\mathbf{W} = \mathbf{V}\Sigma^{-1}\mathbf{U}^T$ performs each linear operation in the reverse order. Diagram adapted from~\textcite{Strang1988}.}	
\label{fig:svd}
\end{figure*}

Divide-and-conquer provides a strategy to solve this problem. Rather than trying to solve for $\mathbf{s}$ and $\mathbf{A}$ simultaneously, we focus on finding $\mathbf{A}$. Furthermore, rather than trying to solve for $\mathbf{A}$ all at once, we solve for $\mathbf{A}$ in a piece-meal fashion by cutting up $\mathbf{A}$ into simpler and more manageable parts.

The {\it singular value decomposition} (SVD) is a linear algebra technique that provides a method for dividing $\mathbf{A}$ into several simpler pieces. For any matrix SVD states that
$$\mathbf{A} = \mathbf{U}\Sigma\mathbf{V}^T.$$
Any matrix is decomposed into three ``simpler'' linear operations: a rotation $\mathbf{V}$, a stretch along the axes $\Sigma$, and a second rotation $\mathbf{U}$. Each matrix in the SVD is ``simpler'' because each matrix contains fewer parameters to infer and each matrix is trivial to invert: $\mathbf{U}$ and $\mathbf{V}$ are rotation matrices (or orthogonal matrices) and $\Sigma$ is a diagonal matrix with real, non-negative values. Figure~\ref{fig:svd} provides a familiar graphical depiction of SVD.

We estimate $\mathbf{A}$ and its inverse $\mathbf{W}$ by recovering each piece of the decomposition individually:
\begin{equation}
\mathbf{W} = \mathbf{A}^{-1} = \mathbf{V}\Sigma^{-1}\mathbf{U}^T
\label{eq:w}
\end{equation}
This equation exploits the fact that the inverse of a rotation matrix is its transpose, (i.e. $\mathbf{V}^{-1} = \mathbf{V}^T$, see Appendix). Furthermore, because $\mathbf{A}$ is invertible by assumption, $\Sigma^{-1}$ exists and is well defined.

The tutorial will proceed by solving for the unmixing matrix $\mathbf{W}$ in two successive stages:
\begin{enumerate}
\item Examine the covariance of the data $\mathbf{x}$ in order to calculate $\mathbf{U}$ and $\Sigma$. (Sections \ref{sec:cov} and \ref{sec:whiten})
\item Return to the assumption of independence of $\mathbf{s}$ to solve for $\mathbf{V}$. (Sections \ref{sec:independence} and \ref{sec:info-theory})
\end{enumerate}
Finally, we present a complete solution to ICA, consider the limits and failures of this technique, and offer concrete suggestions for interpreting results from ICA.

At first encounter the divide-and-conquer strategy might appear weakly motivated but in retrospect, dividing the problem into these parts provides a natural strategy that mirrors the structure of correlations in any data set. We will emphasize these larger points throughout the tutorial to provide the reader a framework for thinking about data in general.

\section{Examining the covariance of the data\label{sec:cov}}

The goal of this section is to explore the covariance of the data given our assumptions, and in the process recover two of the three matrix operations constituting $\mathbf{W}$ (Equation~\ref{eq:w}). The covariance of the data provides an appropriate starting point because the covariance matrix measures all correlations that can be captured by a linear model.\footnote{More specifically, the covariance matrix is a square symmetric matrix where the $ij^{th}$ value is the covariance between $x_i$ and $x_j$. The $ij^{th}$ term measures all second-order correlations that can be captured by a linear model between $x_i$ and $x_j$. For simplicity we assume that the mean of the data is zero or that the mean has been subtracted off for each dimension.}

As a reminder, the covariance is the expected value of the outer product of individual data points $\left<\mathbf{xx}^T\right>$. In order to recover $\Sigma$ and $\mathbf{U}$ we make one additional assumption in a seemingly unmotivated fashion: assume that the covariance of the sources $\mathbf{s}$ is {\em whitened}, or equivalently $\left<\mathbf{ss}^T\right> = \mathbf{I}$, where $\mathbf{I}$ is the identity matrix. We discuss what this assumption means in the following section, but for now we make this assumption blindly and will see what this implies about the observed data $\mathbf{x}$. 

The covariance of the data can be expressed in terms of the underlying sources by plugging in the linear mixture model (Equation~\ref{eq:linear-mixture}):
\begin{eqnarray*}
\left< \mathbf{x} \mathbf{x}^T \right > & = & \left< (\mathbf{A}\mathbf{s}) (\mathbf{A}\mathbf{s})^T \right> \\
							  & = & \left< (\mathbf{U}\Sigma\mathbf{V}^T\mathbf{s}) \; (\mathbf{U}\Sigma\mathbf{V}^T\mathbf{s})^T \right> \\
							  & = & \mathbf{U}\Sigma\mathbf{V}^T   \left<\mathbf{s\;s}^T\right>    \mathbf{V}\Sigma\mathbf{U}^T
\end{eqnarray*}
We exploit our assumption about $\mathbf{s}$ (i.e. $\left<\mathbf{ss}^T\right> = \mathbf{I}$) and a property of orthogonal matrices (i.e. $\mathbf{V}^T=\mathbf{V}^{-1}$) to arrive at a final expression
\begin{equation}
\left< \mathbf{x} \mathbf{x}^T \right > = \mathbf{U}\Sigma^2\mathbf{U}^T.
\label{eq:covariance-x}
\end{equation}
By our shrewd choice of assumption, note that the covariance of the data is independent of sources $\mathbf{s}$ as well as $\mathbf{V}$!

What makes Equation~\ref{eq:covariance-x} extra-special is that it expresses the covariance of the data in terms of a diagonal matrix $\Sigma^2$ sandwiched between two orthogonal matrices $\mathbf{U}$. Hopefully, the form of Equation~\ref{eq:covariance-x} looks familiar to students of linear algebra, but let us make this form explicit.

As an aside, linear algebra tells us that any symmetric matrix (including a covariance matrix) is orthogonally diagonalized by their eigenvectors.\footnote{{\it Orthogonal diagonalization} is a standard decomposition in linear algebra. The term refers to the operation of converting an arbitrary matrix $\mathbf{A}$ into a diagonal matrix by multiplying by an orthogonal basis. Orthogonal diagonalization is achieved by an orthogonal basis of eigenvectors stacked as columns and a diagonal matrix composed of eigenvalues.} Consider a matrix $\mathbf{E}$ whose columns are the eigenvectors of the covariance of $\mathbf{x}$. We can prove that
\begin{equation}
\left<\mathbf{xx}^T \right> = \mathbf{EDE}^T
\label{eq:eigenvectors}
\end{equation}
where $\mathbf{D}$ is a diagonal matrix of associated eigenvalues (see Appendix). The eigenvectors of the covariance of the data form an orthonormal basis meaning that $\mathbf{E}$ is an orthogonal matrix.

Compare Equations~\ref{eq:covariance-x} and~\ref{eq:eigenvectors}. Both equations state that the covariance of the data can be diagonalized by an orthogonal matrix. Equation~\ref{eq:covariance-x} provides a decomposition based on the underlying assumption of ICA. Equation~\ref{eq:eigenvectors} provides a decomposition based purely on the properties of symmetric matrices.

Diagonalizing a symmetric matrix using its eigenvectors is a {\it unique} solution up to a permutation, i.e. no other basis can diagonalize a symmetric matrix. Therefore, if our assumptions behind ICA are correct, then we have identified a partial solution to $\mathbf{A}$: $\mathbf{U}$ is a matrix of the stacked eigenvectors of the covariance of the data and $\Sigma$ is a diagonal matrix with the square root of the associated eigenvalue in the diagonal.

Let us summarize our current state. We are constructing a new matrix $\mathbf{W}$ via Equation~\ref{eq:w}. We have identified the latter two matrices such that 
$$\mathbf{W} = \mathbf{V}\mathbf{D}^{-\frac{1}{2}}\mathbf{E}^{T}.$$
$\mathbf{D}$ and $\mathbf{E}$ are the eigenvalues and eigenvectors of the covariance of the data $\mathbf{x}$. $\mathbf{V}$ is the sole unknown rotation matrix. Let us now take a moment to interpret our results.

\section{Whitening, Revisited\label{sec:whiten}}

\begin{figure}
\centering
\includegraphics[width=0.25\textwidth]{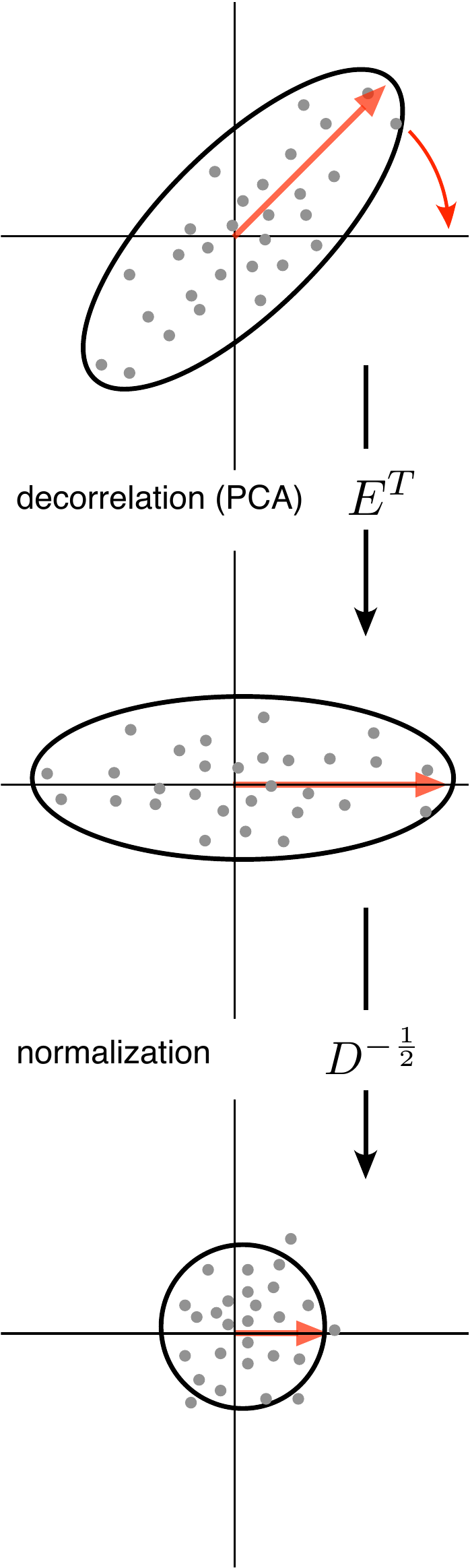}
\caption{Whitening a data set can be represented as a series of two linear operations. Data is projected on the principal components, $\mathbf{E}^T\mathbf{x}$. Each axis is then scaled so that every direction has unit variance, $\mathbf{D}^{-\frac{1}{2}}\mathbf{E}^T\mathbf{x}$. The red arrow indicates the transformation of the eigenvector with the largest variance.}
\label{fig:whiten}
\end{figure}

The solution for ICA thus far performs an operation familiar to signal processing termed {\it whitening}. Whitening is an operation that removes all linear dependencies in a data set (i.e. second-order correlations) and normalizes the variance along all dimensions. Colloquially, this operation is termed {\em sphereing} the data as intuitively, whitening maps the data into a spherically symmetric distribution.

This intuition is demonstrated by the two operations $\mathbf{D}^{-\frac{1}{2}}\mathbf{E}^T$ depicted in Figure~\ref{fig:whiten}. In the first step the data is rotated to align the eigenvectors of the covariance along the cartesian basis. Multiplying by $\mathbf{E}^T$ performs this rotation in order to decorrelate the data, i.e. remove linear dependencies. Mathematically, decorrelation means that the covariance of the transformed data is diagnolized. In our case,
$$\left<(\mathbf{E}^T\mathbf{x})\;(\mathbf{E}^T\mathbf{x})^T\right> = \mathbf{D}$$
where $\mathbf{D}$ is a diagonal matrix of the eigenvalues.\footnote{This equation directly results from orthogonal diagonalization. Starting from Equation~\ref{eq:eigenvectors},\begin{eqnarray*}\left<\mathbf{x}\mathbf{x}^T\right> & = & \mathbf{EDE}^T\\ \mathbf{E}^T\left<\mathbf{x}\mathbf{x}^T\right>\mathbf{E} & =& \mathbf{D}\\ \left<\mathbf{E}^T\mathbf{xx}^T\mathbf{E}\right> & = & \mathbf{D}\\ \left<(\mathbf{E}^T\mathbf{x})\;(\mathbf{E}^T\mathbf{x})^T\right> & = & \mathbf{D}\end{eqnarray*}We have exploited the property that $\mathbf{E}$ is an orthogonal matrix as well as the linearity of expectations.} Each diagonal entry in $\mathbf{D}$ is an eigenvalue of the covariance of the data and measures the variance along each dimension.

This operation has a familiar name - {\it principal component analysis} (PCA). The eigenvectors of the covariance of the data are termed the principal components of the data. Projecting a data set onto the principal components removes linear correlations and provides a strategy for dimensional reduction (by selectively removing dimensions with low variance).

The second operation normalizes the variance in each dimension by multiplying with $\mathbf{D}^{-\frac{1}{2}}$ (Figure~\ref{fig:whiten}). Intuitively, normalization ensures that all dimensions are expressed in standard units. No preferred directions exist and the data is rotationally symmetric -- much like a sphere.

In our problem whitening simplifies the ICA problem down to finding a single rotation matrix $\mathbf{V}$. Let us make this simplification explicit by defining
$$\mathbf{x}_w = (\mathbf{D}^{-\frac{1}{2}}\mathbf{E}^T)\;\mathbf{x}$$
where $\mathbf{x}_w$ is the whitened version of the observed data such that $\left<\mathbf{x}_w\;\mathbf{x}_w^T\right> = \mathbf{I}$. Substituting the above equation into Equations~\ref{eq:ica} and \ref{eq:w} simplifies ICA down to solving $\hat{\mathbf{s}} = \mathbf{V}\mathbf{x}_w$. Note that the problem reduces to finding a rotation matrix $\mathbf{V}$.

The simplified form of ICA provides additional insight into the structure of the recovered sources $\hat{\mathbf{s}}$. Figure~\ref{fig:whiten} highlights that whitened data $\mathbf{x}_w$ is rotationally symmetric, therefore rotated whitened data $\mathbf{\hat{s}}$ must likewise be whitened (remember $\hat{\mathbf{s}} = \mathbf{V}\mathbf{x}_w$). This is consistent with our assumptions about $\mathbf{s}$ in Section~\ref{sec:cov}. Note that this implies that there exists multiple whitening filters including $\mathbf{D}^{-\frac{1}{2}}\mathbf{E}^T$ and $\mathbf{V}(\mathbf{D}^{-\frac{1}{2}}\mathbf{E}^T)$
%(Figure~\ref{fig:zca} shows another notable example).

%\begin{figure}
%\centering
%\includegraphics[width=0.25\textwidth]{ZCA.pdf}
%\caption{Example of another notable whitening filter termed zero-phase component analysis (ZCA). Any matrix $\mathbf{W}$ whitens a data set so long as $\mathbf{W}^T\mathbf{W} = \left<\mathbf{xx}^T\right>^{-1}$ (see Appendix). The linear transform $\mathbf{E}\mathbf{D}^{-\frac{1}{2}}\mathbf{E}^T$ is one such filter that is a symmetric matrix where the final rotation $\mathbf{E}$ ``undoes'' the original PCA rotation. Importantly, ZCA is the whitening filter that minimizes the mean squared error between the original and whitened data $\argmin\mathbf{W}\left<(\mathbf{x} - \mathbf{Wx})^2\right>$. The name ZCA derives from the fact that the eigenvectors of a symmetric (Toeplitz) matrix are sine functions, whose phase is zero. The red arrow indicates the transformation of the eigenvector with the largest variance.}
%\label{fig:zca}
%\end{figure}

\section{The statistics of independence \label{sec:independence}}

The goal of ICA is to find the linear transformation $\mathbf{W}$ that recovers the linear mixed sources $\mathbf{s}$ from the data. By assumption $\mathbf{x} = \mathbf{A}\mathbf{s}$ where $\mathbf{x}$ is the data and both $\mathbf{A}$ and $\mathbf{s}$ are unknown. Exploiting the decomposition of $\mathbf{A}$ and whitening the data has reduced the problem to finding a rotation matrix $\mathbf{V}$ such that $\hat{\mathbf{s}} = \mathbf{V}\mathbf{x}_w$. We now need to exploit the statistics of independence to identify $\mathbf{V}$.

Remember that the covariance matrix measures the linear dependence between all pairs of variables based on second-order correlations. Whitening the data removed all second-order correlations, hence discerning the last rotation matrix $\mathbf{V}$ requires examining other measures of dependency.

Statistical independence is the strongest measure of dependency between random variables. It requires that neither second-order correlations nor higher-order correlations exist. In particular, if two random variables $a$ and $b$ are independent, then $P(a,b) = P(a)\;P(b)$ -- i.e. the joint probability factorizes. In the context of ICA, we assume that all sources are statistically independent, thus
$$P(\mathbf{s}) = \prod_i P(s_i).$$
This implies that the joint distribution of sources $P(\mathbf{s})$ is a special family of distributions termed a {\em factorial distribution} because the joint distribution is the product of the distribution of each source $P(s_i)$.

The problem of ICA searches for the rotation $\mathbf{V}$ such that $\hat{\mathbf{s}}$ is statistically independent $P(\hat{\mathbf{s}}) = \prod_i P(\hat{s}_i)$. Because all second-order correlations have been removed, the unknown matrix $\mathbf{V}$ must instead remove all higher-order correlations. Removing all higher-order correlations with a single rotation $\mathbf{V}$ is a tall order, but if the model $\mathbf{x} = \mathbf{A}\mathbf{s}$ is correct, then this is achievable and $\hat{\mathbf{s}}$ will be statistically independent.

We therefore require a function (termed a {\it contrast function}) that measures the amount of higher-order correlations -- or equivalently, measures how close the estimated sources $\hat{\mathbf{s}}$ are to statistical independence.

\section{A Solution Using Information Theory\label{sec:info-theory}}

The entire goal of this section is to find the last rotation $\mathbf{V}$ such that the estimate of $\hat{\mathbf{s}}$ is statistically independent. To solve this problem, we resort to a branch of mathematics called {\it information theory}. The mathematics of information theory is quite distinct. I would suggest that those new to this material skim this section in their first reading.

Many contrast functions exist for measuring statistical independence of $\hat{\mathbf{s}}$. Examples of contrast functions include rigorous statistical measures of correlations, approximations to said measures (e.g. see code in Appendix~\ref{sec:code}) and clever, ad-hoc guesses.

Because this tutorial is focused on presenting the foundational ideas behind ICA, we focus on a natural measure from information theory to judge how close a distribution is to statistical independence. The {\it mutual information} measures the departure of two variables from statistical independence. The {\it multi-information}, a generalization of mutual information, measures the statistical dependence between multiple variables:
$$I(\mathbf{y}) = \int P(\mathbf{y}) \log_2 \frac{P(\mathbf{y})}{\prod_i P(y_i)} d\mathbf{y}$$
It is a non-negative quantity that reaches a minimum of zero if and only if all variables are statistically independent. For example, if $P(\mathbf{y}) = \prod_i P(y_i)$, then $\log(1) = 0$ and $I(\mathbf{y}) = 0$.

The goal of ICA can now be stated succinctly. Find a rotation matrix $\mathbf{V}$ such that $I(\mathbf{\hat{s}}) = 0$ where $\mathbf{\hat{s}} = \mathbf{V}\mathbf{x}_{w}$. If we find such a rotation, then $\hat{\mathbf{s}}$ is statistically independent. Furthermore, $\mathbf{W} = \mathbf{V}\mathbf{D}^{-\frac{1}{2}}\mathbf{E}^{T}$ is the solution to ICA and we can use this matrix to estimate the underlying sources.

\begin{figure}
\centering
\includegraphics[width=0.98\columnwidth]{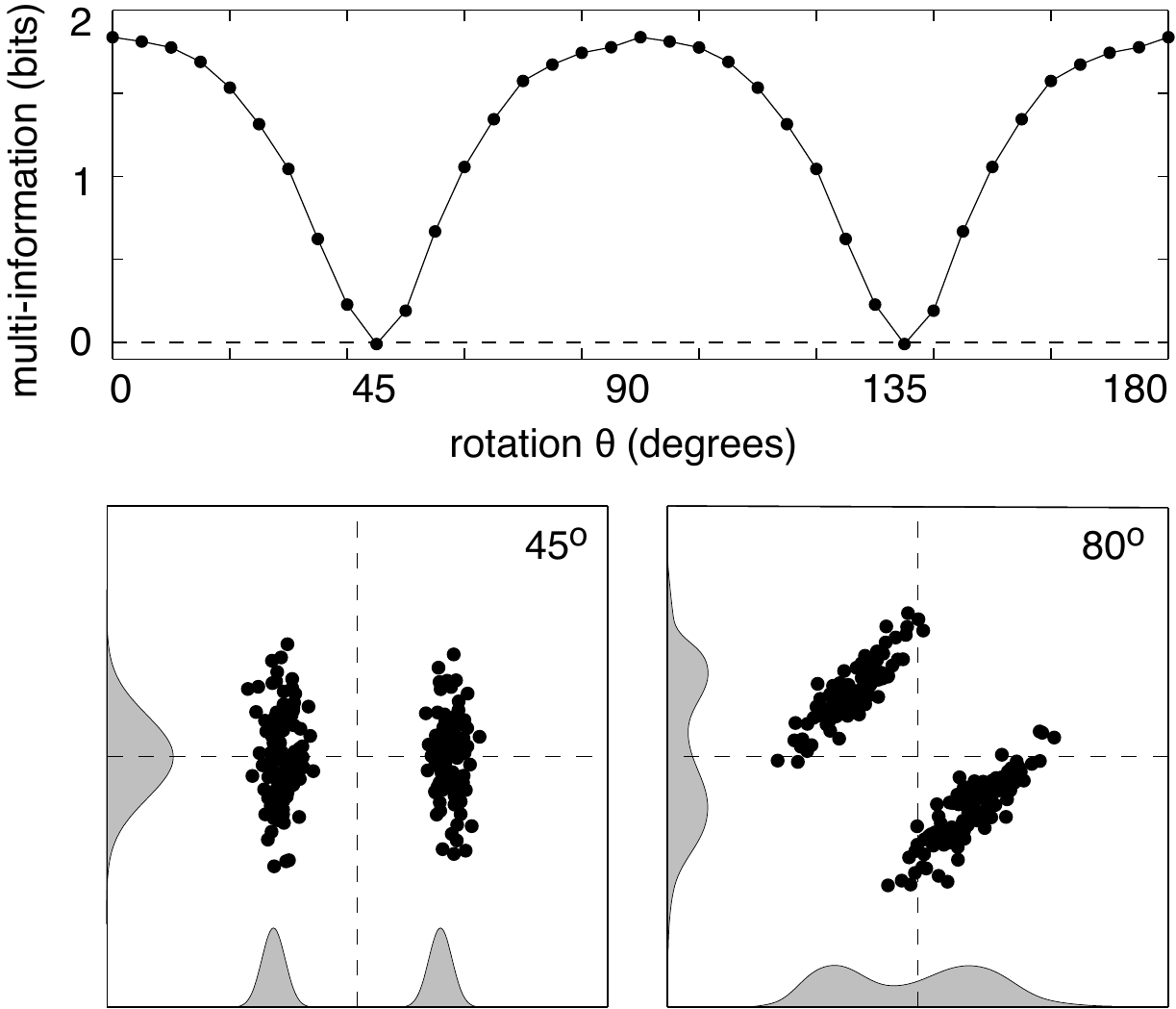}
\caption{Optimization to recover statistical independence for the middle example from Figure~\ref{fig:examples} (see Appendix~\ref{sec:code}). Rotation matrix $\mathbf{V}$ contains one free variable, the rotation angle $\theta$. The multi-information of $\hat{\mathbf{s}}$ is plotted versus the rotation angle $\theta$ (top). Recovered sources $\hat{\mathbf{s}}$ are plotted for two estimates of $\mathbf{W}=\mathbf{VD}^{-\frac{1}{2}}\mathbf{E}^T$ where $\mathbf{V}$ is a rotation matrix with a $45^\circ$ and $80^\circ$ rotation, respectively (bottom).  Grey curves are marginal distributions along each dimension (bottom, inset). Note that a rotation of $45^\circ$ degrees recovers a bimodal and unimodal gaussian distribution along each axis.}
\label{fig:optimization}
\end{figure}

The multi-information is minimized when $\hat{\mathbf{s}}$ is statistically independent, therefore the goal of ICA is to minimize the multi-information until it reaches zero. Reconsider the data from the middle example in Figure~\ref{fig:examples}. $\mathbf{V}$ is a rotation matrix and in two dimensions $\mathbf{V}$ has the form
$$\mathbf{V} = \left[  \begin{array}{rr} \cos(\theta) & -\sin(\theta) \\ \sin(\theta) & \cos(\theta) \end{array}\right].$$
The rotation angle $\theta$ is the only free variable. We can calculate the multi-information of $\hat{\mathbf{s}}$ for all $\theta$ in Figure~\ref{fig:optimization}. Examining the plot we notice that indeed the multi-information is zero when the rotation matrix is $45^\circ$. This means that the recovered distributions are statistically indepndent. The minimum of the multi-information is zero implying that $\mathbf{V}$ is a two-dimensional  $45^\circ$ rotation matrix. 

On the surface the optimization of multi-information might appear abstract. This procedure though can be visualized (and validated) by plotting the recovered sources $\hat{\mathbf{s}}$ using different rotations (Figure~\ref{fig:optimization}, bottom). At a $45^\circ$ rotation the recovered sources $\hat{\mathbf{s}}$ along the x-axis and y-axis are the original bimodal and unimodal gaussian distributions, respectively (Figure~\ref{fig:optimization}, bottom left). Note though that the optimal rotation is not unique as adding integer multiples of $90^\circ$ also minimizes the multi-information!

Minimizing the multi-information is difficult in practice but can be simplified. A simplified form of the optimization bares important relationships with other interpretations of ICA.

The multi-information is a function of the {\it entropy} $H[\cdot]$ of a distribution. The entropy $H\left[\mathbf{y}\right] = -\int{P(\mathbf{y}) \log_2 P(\mathbf{y}) \;d\mathbf{y}}$ measures the amount of uncertainty about a distribution $P(\mathbf{y})$. The multi-information is the difference between the sum of entropies of the marginal distributions and the entropy of the joint distribution, i.e. $I(\mathbf{y}) = \sum_i H\left[y_i\right] - H\left[\mathbf{y}\right]$. Therefore, the multi-information of $\hat{\mathbf{s}}$ is
\begin{eqnarray*}
I(\mathbf{\hat{s}}) & = &  \sum_i H\left[(\,\mathbf{Vx}_{w})_i\,\right] - H\left[\mathbf{Vx}_{w}\right] \\
		 & = & \sum_i H\left[(\,\mathbf{Vx}_{w})_i\,\right] - \left(\;H\left[\mathbf{x}_{w}\right] + \log_2 |\mathbf{V}| \;\right)
\end{eqnarray*}
where $(\mathbf{Vx}_w)_i$ is the $i^{th}$ element of $\hat{\mathbf{s}}$ and we have employed an expression relating the entropy of a probability density under a linear transformation.\footnote{For a linear transformation $\mathbf{B}$ and a random variable $\mathbf{x}$, $H[\mathbf{Bx}] = H[\mathbf{x}] + \log_2 |\mathbf{B}|.$} The determinant of a rotation matrix is 1 so the last term is zero, i.e. $\log_2 |\mathbf{V}| = 0$.

The optimization is simplified further by recognizing that we are only interested in finding the rotation matrix (and not the value of the multi-information).
The term $H[\mathbf{x}_{w}]$ is a constant and independent of $\mathbf{V}$, so it can be dropped:
\begin{equation}
\mathbf{V} = \argmin  \mathbf{V} \;\;\sum_i H\left[(\,\mathbf{Vx}_{w})_i\,\right]
\label{eq:optimization}
\end{equation}
The optimization has simplified to finding a rotation matrix that minimizes the sum of the marginal entropies of $\mathbf{\hat{s}}$. The rotation matrix $\mathbf{V}$ that solves Equation~\ref{eq:optimization} maximizes the statistical independence of $\hat{\mathbf{s}}$.

A few important connections should be mentioned at this point. First, calculating the entropy from a finite data set is difficult and should be approached with abundant caution.\footnote{The entropy is a function of an unknown distribution $P(y)$. Instead any practitioner solely has access to a finite number of samples from this distribution. This distinction is of paramount importance because any estimate of entropy based on finite samples of data can be severely biased. This technical difficulty has motivated new methods for estimating entropy which minimizes such biases (see code in Appendix).} Thus, many ICA optimization strategies focus on approximations to Equation~\ref{eq:optimization} (but see \textcite{Lee03}).

Second, the form of Equation~\ref{eq:optimization} has multiple interpretations reflecting distinct but equivalent interpretations of ICA. In particular, a solution to Equation~\ref{eq:optimization} finds the rotation that maximizes the ``non-Gaussianity'' of the transformed data.\footnote{The deviation of a probability distribution from a Gaussian is commonly measured by the {\it negentropy}. The negentropy is defined as the {\it Kullback-Leibler divergence} of a distribution from a Gaussian distribution with equal variance. Negentropy is equal to a constant minus Equation~\ref{eq:optimization} and maximizing the sum of the marginal negentropy (or non-Gaussianity) is equivalent to minimizing the sum of the marginal entropy.} Likewise, Equation~\ref{eq:optimization} is equivalent to finding the rotation that maximizes the log-likelihood of the observed data under the assumption that the data arose from a statistically independent distribution. Both interpretations provide starting points that other authors have employed to derive Equation~\ref{eq:optimization}.

In summary, we have identified an optimization (Equation~\ref{eq:optimization}) that permits us to estimate $\mathbf{V}$ and in turn, reconstruct the original statistically independent source signals $\hat{\mathbf{s}} = \mathbf{W}\mathbf{x}$. The columns of $\mathbf{W}^{-1}$ are the independent components of the data.

\section{Discussion \label{sec:discussion}}

\begin{figure}
\framebox{\parbox{3.2in}{
{\bf Quick Summary of ICA}
{\sf
\begin{enumerate}
\item Subtract off the mean of the data in each dimension.
\item Whiten the data by calculating the eigenvectors of the covariance of the data.
\item Identify final rotation matrix that optimizes statistical independence (Equation~\ref{eq:optimization}).
\end{enumerate}
}}}
\caption{Summary of steps behind ICA. The first two steps have analytical solutions but the final step must be optimized numerically. See Appendix~\ref{sec:code} for example code.}
\label{fig:summary}
\end{figure}

The goal of ICA is to recover linear mixed signals without knowledge of the linear mixture. ICA recovers the linear mixture and the underlying sources by assuming that the signals are statistically independent. Although a recent addition to the modern toolbox, ICA has found increasing popularity in signal processing and machine learning for filtering signals and performing dimensional reduction (for summary, see Figure~\ref{fig:summary}; for example code, see Appendix~\ref{sec:code}).

Identifying the linear mixture, i.e. the basis of independent components, provides a means for selecting (or deleting) individual sources of interest. For instance, in the cocktail party example in Figure~\ref{fig:cocktail}, each recovered independent component is a column of $\mathbf{W}^{-1}$ and corresponds to the filter selecting the voice or the background music from the pair of microphones. Multiplying the data $\mathbf{x}$ by an individual row of the unmixing matrix $\mathbf{W}$ recovers an estimate of the voice or the music respectively. Note that if one records some new data from the microphones, then the unmixing matrix could be applied to the new data as well.

The success of ICA must be tempered with several computational issues and inherent ambiguities in the solution. In the computation of the unmixing matrix $\mathbf{W} = \mathbf{V}\Sigma^{-1}\mathbf{U}^T$, the matrices  $\Sigma^{-1}$ and $\mathbf{U}$ are analytically calculated from the data. The matrix $\mathbf{V}$ has no analytical form and must be approximated numerically through an optimization procedure. The optimization is inherently difficult because local minima exist in the objective function (Equation~\ref{eq:optimization}) complicating any procedure based on ascending a gradient. In addition, estimating the quantity that must be optimized (the entropy of a distribution) from a finite number of data points is extremely difficult. These two challenges are often approached by approximating the entropy with related statistics (e.g. measures of correlation) that are easier to optimize with finite data sets.

\begin{figure}
\centering
\includegraphics[width=0.98\columnwidth]{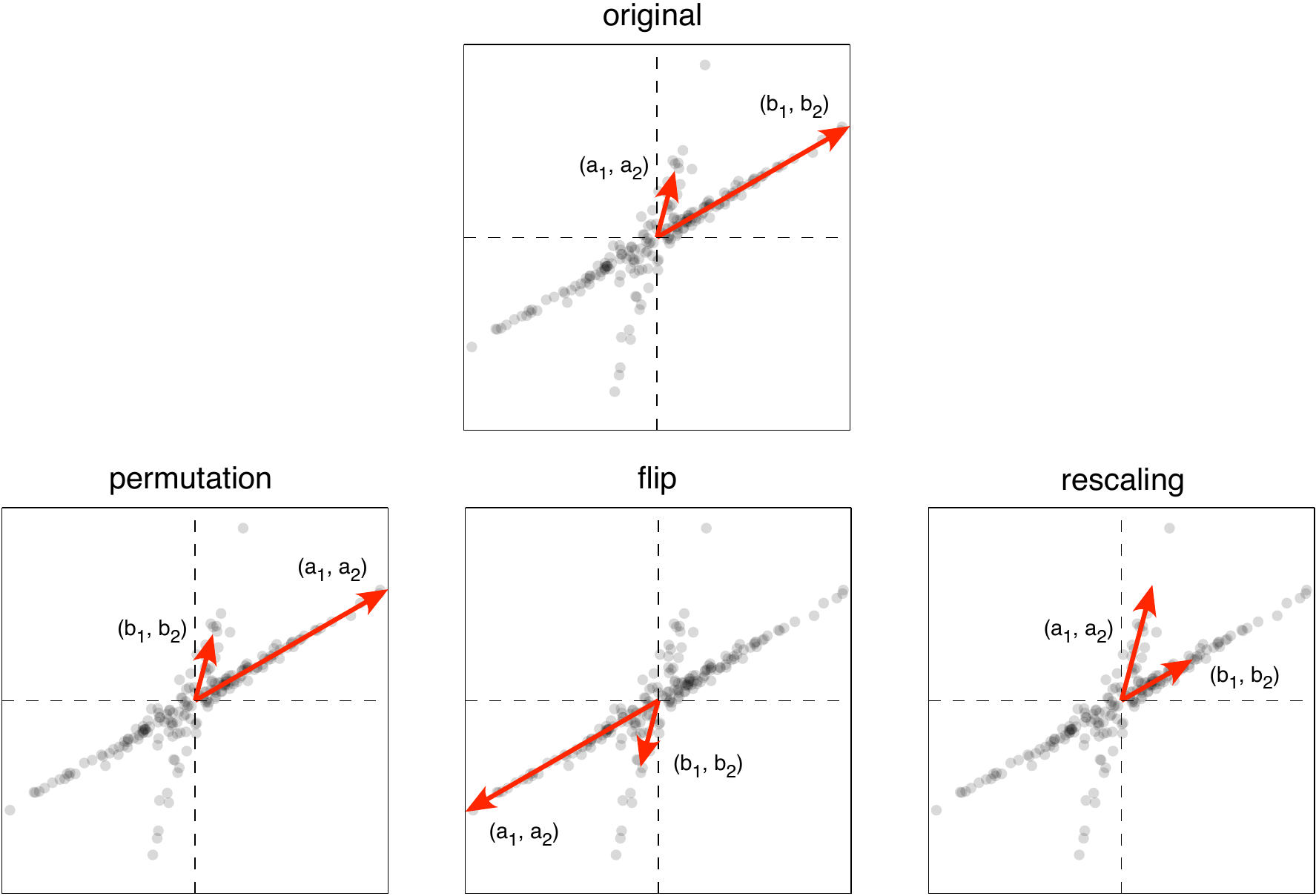}
\caption{Ambiguities in the ICA solution. Each example demonstrates a degree of freedom that provides an additional solution for the top panel (reproduction of Figure~\ref{fig:examples}). {\em Left.} A permutation of the labels of the recovered independent components. {\em Middle.} A flip of the recovered independent components. {\em Right.} A rescaling of the recovered independent components.}
\label{fig:ambiguities}
\end{figure}

Even when the above issues are addressed, any ICA solution is subject to several inherent ambiguities. These ambiguities exist because the objective is agnostic to these degrees of freedom (Figure~\ref{fig:ambiguities}). First, the labels of each independent component can be arbitrarily permuted.\footnote{Since $\mathbf{W}$ undoes the linear transformation of $\mathbf{A}$, we might infer that $\mathbf{WA} = \mathbf{I}$, where $\mathbf{I}$ is an identity matrix. This is overly restrictive, however and any permutation of the identity matrix is also a valid ICA solution, e.g. $\mathbf{WA} = \left[  \begin{array}{rr} 1 & 0 \\ 0 & 1 \end{array}\right]$ or $\left[  \begin{array}{rr} 0 & 1 \\ 1 & 0 \end{array}\right]$.} Second, any independent component can be flipped across the origin. Third, any independent component can be rescaled with arbitrary length (because one can absorb any rescaling into the inferred sources $\hat{\mathbf{s}}$).\footnote{We resolve the last ambiguity by selecting an arbitrary scale for the recovered sources, in particular a scale that simplifies the computation. For instance, earlier in the text we assumed that the covariance of the source is white $\left<\mathbf{s\;s}^T\right> = \mathbf{I}$. This assumption is a trick that accomplishes two simultaneous goals. First, a variance of 1 in each dimension provides a fixed scale for $\mathbf{s}$. Second, a white covariance simplifies the evaluation of the data covariance by removing any dependence on $\mathbf{s}$ and $\mathbf{V}$ (Equation~\ref{eq:covariance-x}).} For most applications these ambiguities are inconsequential, however being cognizant of these issues could prevent an inappropriate interpretation of the results.

Given the challenges and ambiguities in the ICA solution highlighted above, one might question what benefit exists in analyzing a data set with ICA. The primary advantage of ICA is that it provides a fairly unrestrictive linear basis for representing a data set. This advantage is often highlighted by contrasting the results of ICA with PCA. PCA identifies an orthogonal linear basis which maximizes the variance of the data. ICA is not restricted to an orthogonal basis because statistical independence makes no such requirement. This is most apparent in the top example of Figure~\ref{fig:examples} where the directions of interest are quite salient, but maximizing the explained variance fails to identify these directions. In terms of real world data, this lack of restriction means that independent sources need not be orthogonal but merely linear independent.

These statements are only part of the story as evident in the middle and bottom examples of Figure~\ref{fig:examples}. In both examples the underlying sources are orthogonal. If orthogonality were the only relevant factor, PCA should have recovered the underlying sources in the middle example. Why does ICA uniquely recover the sources in the middle example? The answer lies in the fact that PCA can identify the underlying sources only when data is distributed appropriately, such as a Gaussian distribution. This statement will be justified in the following section along with an analogous statement for ICA. Importantly, this discussion will lead to conditions under which ICA will succeed or fail to recover the underlying sources.

\subsection{What data works and why\label{sec:limits}}

Predicting the success of any dimensional reduction technique ultimately requires a discussion about how the data was generated. The discussion draws heavily on the topic of whitening in Section~\ref{sec:whiten}, although the mathematics are slightly more involved.

Let us begin by examining our assumptions closer. We assumed that the underlying sources are statistically independent, i.e. $P(\mathbf{s}) = \prod_i{\;P(s_i)}$. As mentioned in Section~\ref{sec:whiten}, distributions of this form are termed factorial distributions. Furthermore, the linear mixture model $\mathbf{x} = \mathbf{A}\mathbf{s}$ provides a concrete prescription for how the observed data $\mathbf{x}$ was generated. By assumption ICA expects the data to arise from an (invertible) linear transformation applied to a factorial distribution. Hence, ICA expects and will work optimally on data arising from {\em linearly transformed factorial distributions} -- applying ICA to any other type of data is a gamble with no guarantees.\footnote{In general, it is difficult to write down a succinct, closed-form equation for a linear transformed factorial distribution. The Gaussian distribution is an exception to this statement (see below). In general, the marginal distribution of a linear transformed factorial distribution is the linear sum of two independent distributions. The linear sum of two independent distributions is mathematically equal to the {\em convolution} of the independent distributions with weights given by the coefficients of the independent components.}

The case of Gaussian distributed data provides an important didactic point about the application of ICA (Figure~\ref{fig:examples}, bottom example). Why is principal component analysis (PCA) identical to independent component analysis for this data?

As a side bar, remember that the first part of the ICA solution employs PCA to remove second-order correlations (Section~\ref{sec:whiten}). The second part of the ICA solution employs the rotation $\mathbf{V}$ to remove higher-order correlations (Section~\ref{sec:independence}).

Since PCA is equivalent to ICA for Gaussian data, we infer that Gaussian data contains no higher order correlations -- but we need not rely on inference to understand this part. If the observed data $\mathbf{x}$ is Gaussian distributed with covariance $\Sigma$, applying a whitening filter results in whitened data $\mathbf{x}_w$ Gaussian distributed with no covariance structure (i.e. identity covariance matrix). Note that $P(\mathbf{x}_w)$ can be written as a factorial distribution $P(\mathbf{x}_w) = \prod_i \exp(x_{w,i})$. Hence, a whitening filter based on PCA already achieves a factorial distribution and no need exists for computing $\mathbf{V}$. 

This discussion has been theoretical, but it has important implications for the application of ICA on real data. Often, a practitioner of ICA does not know how their data arose and merely tries this technique to see if the results are reasonable. One lesson from this discussion is to ask how prominent higher-order correlations are in the data. If higher-order correlations are small or insignificant, then PCA is sufficient to recover the independent sources of the data.

A second lesson is to ask how independent the recovered sources are. Empirically measuring $I(\hat{\mathbf{s}})$ (or any measure of correlation) provides a useful check to ensure that the sources are indeed independent. In practice, $I(\hat{\mathbf{s}})$ rarely achieves $0$ either because of sampling issues (see Appendix) or because of local minima in the optimization. Therefore, the final results must be rigorously checked by calculating multiple measures of correlation or employing bootstrap procedures.

\subsection{Extensions}

\begin{figure}
\centering
\includegraphics[width=0.98\columnwidth]{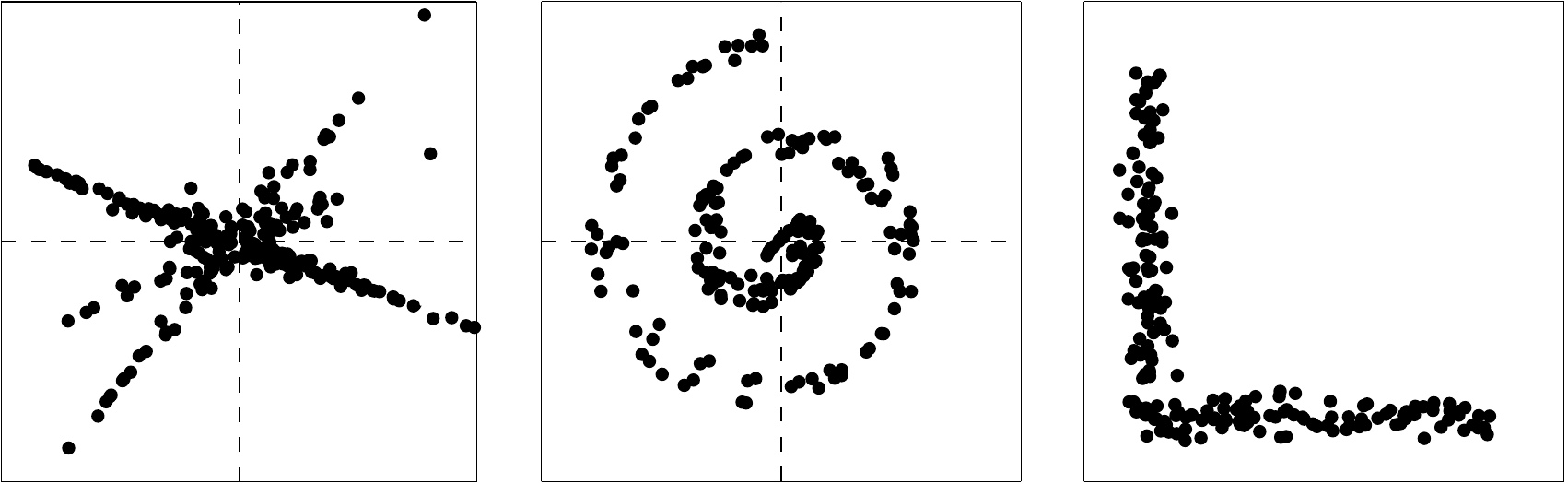}
\caption{Challenges to identifying independent components. {\em Left}: Overcomplete representation. The number of independent sources exceed the number of measurements. {\em Middle}: Nonlinear manifold. The underlying source lies on a nonlinear manifold. {\em Right}: Occlusion. The superposition of the two sources is not linear but mutually exclusive.}
\label{fig:challenges}
\end{figure}

The tutorial up to this point has offered a foundation for approaching the topic of ICA. A judicious reader might have already recognized several deficiencies in the ICA framework which could readily be improved or extended. The purpose of this section is  to highlight a few future directions that a reader would now be prepared to explore. 

An obvious shortcoming of ICA can be seen as far back as Figure~\ref{fig:cocktail}. In retrospect, it seems artificial to stipulate two microphones for separating the two auditory sources. Clearly, a single microphone should suffice for discerning between a potentially large number of auditory sources. Mathematically, this intuition suggests that the number of sources need not match the number of recordings. In particular, the dimension of $\mathbf{s}$ might exceed the dimension of $\mathbf{x}$. This {\em overcomplete} problem is a situation in which $\mathbf{A}$ is not a square matrix (Figure~\ref{fig:challenges}, left). Therefore, $\mathbf{A}$ is not an invertible matrix, violating an important assumption behind the discussion of ICA.

If $\mathbf{A}$ is not invertible, then outside information must be employed to solve the problem. Additional information often takes the form of adding a regularization term, or positing a particular distribution for the data. The latter technique is a Bayesian perspective in which one posits that the data arose from a {\em prior distribution} and the observations are corrupted by some noise process stipulated by a likelihood function. Several versions of ICA have been developed that follow this Bayesian perspective and provide solutions (although limited) to various over-complete situations.

Multiple extensions have further been suggested for ICA based on empirical observations of blind source separation problems in the real world. For instance, some recent extensions have focused on handling data arising from binary or sparse sources, nonlinear interactions (e.g. occlusion) or nonlinear representations. For those who wish to delve into this topic further, a nice starting place are the popular {\em Infomax} papers \cite{Bell95, Bell97}, a special journal issue on ICA~\cite{Lee03} and an entire text book on this topic~\cite{Hyvarinen2004}. \textcite{LyuSimoncelli09} provide a tremendous discussion on the statistics of ICA and its relationship to natural signal statistics. Finally, \textcite{FaivishevskyGoldberger08} provide an elegant implementation mirroring this tutorial's explanations that achieves state-of-the-art performance.

Writing this paper has been an extremely instructional experience for me. I hope that this paper helps to demystify the motivation and results of ICA, and the underlying assumptions behind this important analysis technique. Please send me a note if this has been useful to you as it inspires me to keep writing! 

\bibliography{ica}

\appendix
\section{Mathematical Proofs}

{\bf 1. The inverse of an orthogonal matrix is its transpose.} \\

Let $\mathbf{A}$ be an $m \times n$ orthogonal matrix where $\mathbf{a_i}$ is th
e $i^{th}$ column vector. The $ij^{th}$ element of $\mathbf{A}^{T}\mathbf{A}$ is
 
$$(\mathbf{A}^{T}\mathbf{A})_{ij} = \mathbf{a_{i}}^{T}\mathbf{a_{j}} =  \left\{ 
	\begin{array}{ll}
		1 & if \;\;i=j \\
		0 & otherwise \\
	\end{array} \right.$$
Therefore, because $\mathbf{A}^{T}\mathbf{A}=\mathbf{I}$, it follows that $\mathbf{A}^{-1}=\mathbf{A}^{T}$. \\

{\bf 2. A symmetric matrix is diagonalized by a matrix of its orthonormal eigenvectors.} \\

Let $\mathbf{A}$ be a square \mbox{$n \times n$} symmetric matrix with associated eigenvectors $\{\mathbf{e_1, e_2, \ldots, e_n} \}$. Let $ \mathbf{E}=[\mathbf{e_1\;e_2\;\ldots\;e_n]}$ where the $i^{th}$ column of $\mathbf{E}$ is the eigenvector $\mathbf{e_i}$. This theorem asserts that there exists a diagonal matrix $\mathbf{D}$ such that $\mathbf{A}=\mathbf{EDE}^{T}$.

This proof is in two parts. In the first part, we see that the any matrix can be orthogonally diagonalized if and only if that matrix's eigenvectors are all linear independent. In the second part of the proof, we see that a symmetric matrix has the special property that all of its eigenvectors are not just linear independent but also orthogonal, thus completing our proof.

In the first part of the proof, let $\mathbf{A}$ be just some matrix, not necessarily symmetric, and let it have independent eigenvectors (i.e. no degeneracy). Furthermore, let $ \mathbf{E}=[\mathbf{e_1\;e_2\;\ldots\;e_n]}$ be the matrix of eigenvectors placed in the columns. Let $\mathbf{D}$ be a diagonal matrix where the $i^{th}$ eigenvalue is placed in the $ii^{th}$ position.

We will now show that $\mathbf{AE=ED}$. We can examine the columns of the right-hand and left-hand sides of the equation.
\begin{displaymath}
\begin{array}{rrcl}
	\mathsf{Left\;hand\;side:} & \mathbf{AE} & = & [\mathbf{Ae_1}\;\mathbf{Ae_2}\;\ldots\;\mathbf{Ae_n}] \\
	\mathsf{Right\;hand\;side:} & \mathbf{ED} & = & [\lambda_{1}\mathbf{e_1}\:\lambda_{2}\mathbf{e_2}\:\ldots\:\lambda_{n}\mathbf{e_n}]
\end{array}
\end{displaymath}
Evidently, if $\mathbf{AE=ED}$ then $\mathbf{Ae_i}=\lambda_{i}\mathbf{e_i}$ for all $i$. This equation is the definition of the eigenvalue equation. Therefore, it must be that $\mathbf{AE=ED}$. A little rearrangement provides $\mathbf{A=EDE}^{-1}$, completing the first part the proof.

For the second part of the proof, we show that a symmetric matrix always has orthogonal eigenvectors. For some symmetric matrix, let $\lambda_{1}$ and $\lambda_{2}$ be distinct eigenvalues for eigenvectors $\mathbf{e_1}$ and $\mathbf{e_2}$.
\begin{eqnarray*}
	\lambda_1\mathbf{e_1}\cdot\mathbf{e_2} & = & (\lambda_1 \mathbf{e_1})^{T} \mathbf{e_2} \\
	& = & (\mathbf{Ae_1})^{T} \mathbf{e_2} \\
	& = & \mathbf{e_1}^{T}\mathbf{A}^{T} \mathbf{e_2} \\
	& = & \mathbf{e_1}^{T}\mathbf{A} \mathbf{e_2} \\
	& = & \mathbf{e_1}^{T} (\lambda_{2}\mathbf{e_2}) \\
	\lambda_1\mathbf{e_1}\cdot\mathbf{e_2} & = & \lambda_2\mathbf{e_1}\cdot\mathbf{e_2}
\end{eqnarray*}
By the last relation we can equate that \mbox{$ (\lambda_1-\lambda_2)\mathbf{e_1}\cdot\mathbf{e_2} = 0$}. Since we have conjectured that the eigenvalues are in fact unique, it must be the case that $\mathbf{e_1}\cdot\mathbf{e_2} = 0$. Therefore, the eigenvectors of a symmetric matrix are orthogonal.

Let us back up now to our original postulate that $\mathbf{A}$ is a symmetric matrix. By the second part of the proof, we know that the eigenvectors of $\mathbf{A}$ are all orthonormal (we choose the eigenvectors to be normalized). This means that $\mathbf{E}$ is an orthogonal matrix so by theorem 1, $\mathbf{E}^T=\mathbf{E}^{-1}$ and we can rewrite the final result.
$$\mathbf{A=EDE}^T$$
Thus, a symmetric matrix is diagonalized by a matrix of its eigenvectors.\\

{\bf 3. For any zero-mean, random vector $\mathbf{x}$, the matrix $\mathbf{W}$ whitens $\mathbf{x}$ if and only if $\mathbf{W}^T\mathbf{W} = \left<\mathbf{xx}^T\right>^{-1}$.} \\

The definition of whitening $\mathbf{x}$ is that the covariance of $\mathbf{Wx}$ is the identity matrix for some matrix $\mathbf{W}$. We begin with this definition and observe what this requires of $\mathbf{W}$.
\begin{eqnarray*}
\left<(\mathbf{Wx}) (\mathbf{Wx})^T \right> & = & \mathbf{I} \\
\mathbf{W} \left<\mathbf{x}\mathbf{x}^T \right> \mathbf{W}^T& = & \mathbf{I} \\
\left<\mathbf{x}\mathbf{x}^T \right> & = & \mathbf{W}^{-1}\mathbf{W}^{-T} \\
\left<\mathbf{x}\mathbf{x}^T \right> ^{-1} & = & \mathbf{W}^{T}\mathbf{W}
\end{eqnarray*}
The left-hand side is the inverse of the covariance of the data $\mathbf{x}$, also known as the precision matrix. Note that the matrix $\mathbf{W}$ contains $n^2$ unknown terms but the equation provides fewer constraints. Multiple solutions for $\mathbf{W}$ must exist in order to whiten the data $\mathbf{x}$.

\section{Code\label{sec:code}}

This code is written for Matlab 7.4 (R2007a). Several popular packages exist for computing ICA including  {\tt FastICA} and {\tt InfoMax} which employs several notable algorithmic improvements for speed and performance.

For the purposes of this tutorial, I am including an implementation of a simple ICAalgorithm that attempts to find the direction of maximum kurtosis \cite{Cardoso89}. The FOBI algorithm is not very robust algorithm although it is quite elegant, simple to code and useful as a didactic tool.

{\footnotesize \input{ica.m}}

The next section of code is a script to demonstrate the example from Figure~\ref{fig:optimization}.

{\footnotesize \input{example.m}}

This function demonstrates a clever binless estimator of entropy which exhibits good statistical properties \cite{Victor02}.

{\footnotesize \input{entropy.m}}

\end{document}